\documentclass{article}

\usepackage{microtype}
\usepackage{graphicx}
\usepackage{subfigure}
\usepackage{booktabs} 

\usepackage{hyperref}

\usepackage[accepted]{icml_arxiv}

\usepackage{amssymb}
\usepackage{mathtools}
\usepackage{amsthm}
\usepackage{multirow}
\usepackage{subcaption}
\usepackage{xcolor}
\usepackage{pifont}
\usepackage{listings}

\usepackage[skins]{tcolorbox}
\usepackage[detect-all]{siunitx}

\long\def\comment#1{}

\usepackage[textsize=tiny]{todonotes}

\definecolor{darkgreen}{RGB}{50,100,0}
\definecolor{darkred}{RGB}{200, 0, 0}
\definecolor{lightred}{RGB}{250, 200, 200}
\definecolor{lightblue}{RGB}{210, 220, 250}

\newcommand{\cmark}{\textcolor{darkgreen}{\ding{51}}} %
\newcommand{\xmark}{\textcolor{darkred}{\ding{55}}} %

\definecolor{rationale}{RGB}{255, 126, 121}
\definecolor{program}{RGB}{91, 155, 213}

\definecolor{mydarkblue}{rgb}{0,0.08,0.45}
\definecolor{mydarkgreen}{RGB}{0, 139, 69}
\definecolor{MAEblue}{RGB}{47 112 182}
\definecolor{SDEblue}{RGB}{28 58 88}
\hypersetup{
	colorlinks=true,
	urlcolor=magenta,
	citecolor=SDEblue,
}
\definecolor{mycyan}{cmyk}{.3,0,0,0}

\icmltitlerunning{Embedding Self-Correction as an Inherent Ability in Large Language Models for Enhanced Mathematical Reasoning}

\begin{document}

\twocolumn[
\icmltitle{Embedding Self-Correction as an Inherent Ability in Large Language Models\\ for Enhanced Mathematical Reasoning}






\icmlsetsymbol{equal}{*}
\icmlsetsymbol{cor}{$\dagger$}

\begin{icmlauthorlist}
\icmlauthor{Kuofeng Gao}{equal,yyy}
\icmlauthor{Huanqia Cai}{equal,yyy}
\icmlauthor{Qingyao Shuai}{yyy}
\icmlauthor{Dihong Gong}{yyy}
\icmlauthor{Zhifeng Li}{cor,yyy}

\end{icmlauthorlist}

\icmlaffiliation{yyy}{Tencent, Shenzhen, Guangdong, China}

\icmlcorrespondingauthor{Zhifeng Li}{zhifeng0.li@gmail.com}

\icmlkeywords{Machine Learning, ICML}

\vskip 0.3in
]



\printAffiliationsAndNotice{\icmlEqualContribution} 

\begin{abstract}
Accurate mathematical reasoning with Large Language Models (LLMs) is crucial in revolutionizing domains that heavily rely on such reasoning. However, LLMs often encounter difficulties in certain aspects of mathematical reasoning, leading to flawed reasoning and erroneous results. To mitigate these issues, we introduce a novel mechanism, the Chain of Self-Correction (CoSC), specifically designed to embed self-correction as an inherent ability in LLMs, enabling them to validate and rectify their own results. The CoSC mechanism operates through a sequence of self-correction stages. In each stage, the LLMs generate a program to address a given problem, execute this program using program-based tools to obtain an output, subsequently verify this output. Based on the verification, the LLMs either proceed to the next correction stage or finalize the answer. This iterative self-correction process allows the LLMs to refine its reasoning steps and improve the accuracy of its mathematical reasoning. 
We implement CoSC using a two-phase fine-tuning approach. First, LLMs are trained with a relatively small volume of seeding data generated from GPT-4. Then, we enhance CoSC by training with a larger volume of self-generated data, without relying on GPT-4.
Experiments show that CoSC significantly boosts performance on standard mathematical datasets compared to existing open-source LLMs. Notably, our CoSC-Code-34B model achieved a 53.5\% score on the challenging MATH dataset, outperforming models like ChatGPT, GPT-4, and multi-modal LLMs such as GPT-4V and Gemini-1.0. Importantly, CoSC operates in a zero-shot manner without requiring demonstrations. 
\end{abstract}

\section{Introduction}
\label{sec: intro}



Large Language Models (LLMs), such as GPT-4 \citep{openai2023gpt4}, have recently demonstrated state-of-the-art performance across a variety of natural language processing (NLP) tasks, including natural language generation and understanding \citep{chowdhery2023palm,team2023gemini,anil2023palm,penedo2023refinedweb,gao2024denial,gao2024benchmarking}. Despite their success, LLMs often struggle with mathematical reasoning tasks due to their lack of explicit logical reasoning and judgment, which are crucial for solving such problems. Moreover, there is a fundamental gap between natural language and the language of mathematical formulas, which further complicates these tasks. As a result, accurate mathematical reasoning remains an essential yet challenging capability for LLMs to develop, in order to further advance various domains. Consequently, it is still an open challenge to tackle mathematical problems for existing open-source LLMs \citep{touvron2023llama}.

\begin{figure*}[t] \centering     
\includegraphics[width=\linewidth]{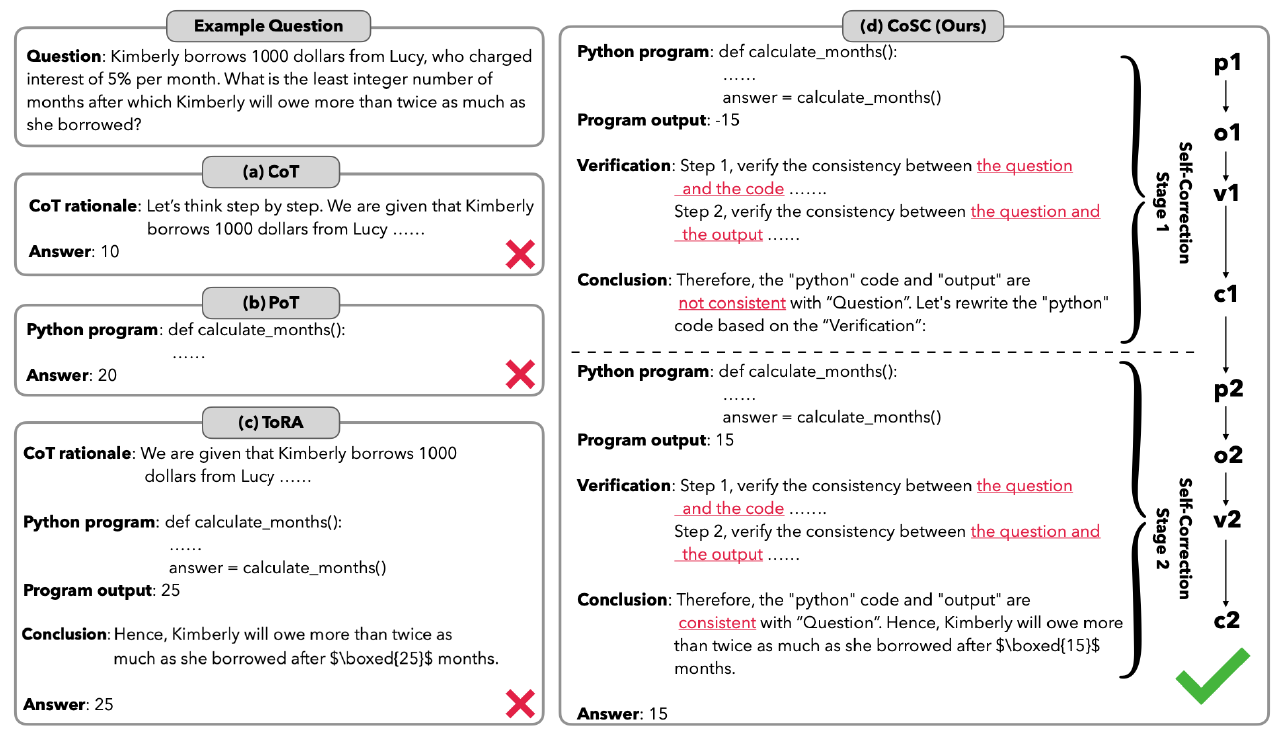} 
\vspace{-2em}
\caption{Comparison of four reasoning frameworks for solving an example mathematical question. (a) Chain of Thoughts (CoT) \citep{wei2022chain}. (b) Program of Thoughts (PoT) \citep{chen2022program}. (c) ToRA \citep{gou2023tora} that incorporates CoT, PoT, and the utilization of tools. (d) Our proposed CoSC consists of a sequence of multiple self-correction stages (two stages are shown in this example). Each stage has four sub-stages: (\textbf{p1}) LLMs generate program \textit{w.r.t.} the question; (\textbf{o1}) execute the program to obtain program output; (\textbf{v1}) perform two-step verification for consistency of the question with both the generated program and the program output; (\textbf{c1}) conclude a refined answer or continue the next subsequent self-correction stage depending on the verification result. The final answer is extracted from the last conclusion sub-stage with regular expression matching.}
\vspace{-1em}
\label{example}
\end{figure*}

To improve the mathematical reasoning abilities, numerous approaches have been investigated in previous research, including prompting \citep{wei2022chain,chen2022program,wang2022self}, pretraining \citep{azerbayev2023llemma,fu2023kwaiyiimath,shao2024deepseekmath}, and finetuning \citep{luo2023wizardmath,yu2023metamath,yue2023mammoth,gou2023tora,liu2023improving}. In particular, finetuning has become a favored technique among them, which updates open-sourced LLMs based on previously generated high-quality question-response pair datasets. Compared to open-source LLMs, finetuning has demonstrated significant improvement, but there is still potential room for further enhancement.



Most current methods \citep{yu2023metamath,gou2023tora} generate finetuning datasets by prompting GPT-4 to rephrase mathematical questions from various perspectives or incorporate chain-of-thoughts (CoT) analysis \citep{wei2022chain} and program-of-thoughts (PoT) code \citep{chen2022program} to diversify mathematical responses. As a result, LLMs trained on these datasets can comprehend different questions
and learn to use code to solve mathematical problems. However, precise multi-round reasoning capabilities remain challenging for them.  
Once potential errors occur throughout reasoning stages, it can lead to incorrect results. Consequently, it is essential to incorporate a self-correction mechanism into mathematical responses, which can enable LLMs to learn to correct themselves in multiple rounds. 
Besides, while supervised finetuning datasets can be developed with GPT-4's assistance, it still requires human experiments with different prompts and the cost of using the interface. This highlights the need for effective finetuning with unlabeled datasets.


To address the aforementioned challenges, our study introduces the Chain of Self-Correction (CoSC), a novel mechanism designed to embed self-correction as an inherent capability in LLMs, enabling them to validate and rectify their own results. The CoSC mechanism operates through a sequence of self-correction stages, where LLMs generate a program to solve a given problem, execute the program using program-based tools to obtain an output, and subsequently verify this output. Depending on the verification, the LLMs either advance to a subsequent stage of self-correction or conclude with the refined solution. An example of our CoSC reasoning trajectory with multiple
self-correction stages (two stages are shown in this example) is shown in Fig. \ref{example}.  This iterative self-correction process allows the LLMs to refine their reasoning steps and improve the accuracy of its mathematical reasoning. 

To implement the CoSC mechanism at a low cost, we adopt a two-phase finetuning approach. In the first phase, termed the \textbf{CoSC foundational learning}, LLMs are trained with a relatvely small volume of seeding data generated from GPT-4, equipping them with a baseline proficiency in the CoSC methodology. In particular, we prompt GPT-4 with training questions from  MATH \citep{hendrycks2021measuring} and GSM8K \citep{cobbe2021training} datasets to generate mathematical reasoning trajectories that adhere to the CoSC protocol. Specifically, each generated trajectory consists of program-of-thoughts code, program output, a two-step verification process that ensures the alignment of the question with both the generated program and the resulting output, and a conclusion to determine whether the trajectory should be refined or if the final answer can be provided.

Using GPT-4 can be expensive, especially when dealing with large volumes of training data. Alternatively, we propose a cost-free method to further boost performance through self enhancement in the second phase. The second phase, referred to as \textbf{CoSC self enhancement}, builds upon foundational learning by further adapting the LLMs obtained from the first phase with self-generated trajectories. These trajectories are produced by the models trained in the foundational phase, allowing for the generation of a substantial volume of data without GPT-4 intervention. In both phases, we retain only the trajectories whose answers match the ground-truth labels of the corresponding questions.

In summary, our study makes the following contributions:
\begin{itemize}
\vspace{-1em}
\item We propose the Chain of Self-Correction (CoSC) mechanism, which effectively embeds self-correction as an inherent ability in Large Language Models (LLMs). Once LLMs learn this ability during training, they can self-correct in a zero-shot setting during inference without the need for external feedback or few-shot demonstrations. With the inherent self-correction ability, even an originally weak LLM is able to achieve excellent performance in mathematical reasoning, as strongly supported by our experimental results. This unique contribution distinguishes our work from related works discussed in Section \ref{sec:Existing_methods_related_to_Self-Correction_in_LLMs} of this field. 
\vspace{-0.5em}

\item To implement the CoSC mechanism at a low cost, we introduce a two-phase finetuning approach. The first phase involves CoSC foundational learning, where we use a relatively small volume of seeding data generated by GPT-4. In the second phase, CoSC self enhancement occurs using a larger volume of self-generated data with the model obtained from the first phase, without relying on paid GPT-4.
\vspace{-0.5em}

\item Our comprehensive experiments
demonstrate that the CoSC mechanism provides a new benchmark for performance on established mathematical datasets when compared to existing open-source LLMs. Notably, our CoSC-Code-34B model achieves superior performance over both closed-source LLMs and some of multi-modal LLMs, particularly on the challenging MATH dataset. 
\vspace{-0.5em}

\item The proposed CoSC mechanism, by embedding self-correction as an inherent capability in LLMs, enables them to think before responding to a question, creating an internal chain of self-correction to progressively verify and rectify their original answers. It is more akin to the slow thinking process of humans, which is particularly helpful in solving difficult mathematical reasoning problems. This approach can provide valuable insights for future research and contribute to the ongoing advancement of LLMs.  
\end{itemize}

\section{Related Work}
\label{sec: relw}

\subsection{LLMs for Mathematical Reasoning  }
\label{sec:LLM for math reasoning}
Mathematical reasoning \citep{liu2023tinygsm,wang2023math,huang2024key,toshniwal2024openmathinstruct,chen2024masked,zhang2024mario} is a challenging reasoning task for LLMs, which requires the ability to understand mathematical concepts, computation and multi-round reasoning. Existing mathematical reasoning approaches can be broadly classified into three categories: (1) \textit{Prompting} methods \citep{wei2022chain,chen2022program,wang2022self} focus on extracting the inherent mathematical reasoning skills of LLMs by utilizing well-crafted prompting strategies during inference. Notably, they leverage the existing knowledge in LLMs without the need for parameter updates. (2) \textit{Pretraining} methods \citep{azerbayev2023llemma,fu2023kwaiyiimath,yang2024qwen2} pre-train LLMs on large-scale corpora containing mathematical problems and related content with language modeling objectives. The goal is to train a base foundation language model as a platform for mathematical domain. (3) \textit{Finetuning} methods \citep{luo2023wizardmath,yu2023metamath,yue2023mammoth,gou2023tora,liu2023improving,gao2024efficient,mitra2024orca} refine the mathematical reasoning capabilities of LLMs by offering more targeted training.
Central to this approach is the generation of high-quality question-response pair datasets, which needs the assistance of complexity-based chain-of-thoughts prompting or tools-based augmentation. 
In this paper, we propose a Chain of Self-Correction (CoSC) along this direction that incorporates an iterative self-correction process into datasets generation.

\subsection{Existing methods related to Self-Correction in LLMs}
\label{sec:Existing_methods_related_to_Self-Correction_in_LLMs}

Mathematical reasoning poses a significant challenge due to its demand for precise multi-round logical reasoning to solve problems. The potential for errors increases with each reasoning step, making it crucial for LLMs to have the ability to self-correct in order to produce accurate results. There are some recent studies \citep{chen2023teaching,gou2023critic,lightman2023let,huang2023large,chen2024boosting} attempt to enable large language models to perform self-correction by either prompting methods or fine-tuning methods. For prompting methods, they can correct their responses when interacting with external tools \citep{gou2023critic,chen2023teaching}, such as search engines and calculators, or designing complex prompts \citep{chen2024boosting}. Notably, prompting methods can be orthogonally combined with finetuning methods. For finetuning methods, previous works \citep{yu2023teaching,an2023learning} only model single-round correction during the training stage, and perform verification in a straightforward manner during the inference stage.

\subsection{Difference between the proposed method and existing self-correction techniques}


The existing self-correction research can be broadly classified into two categories: prompt-based methods \citep{gou2023critic,chen2023teaching,chen2024boosting} and SFT-based methods \citep{yu2023teaching,an2023learning}. Our approach falls under the SFT-based category and fundamentally differs from prompt-based methods. Specifically, prompt-based methods leverage the intrinsic capabilities of LLMs for mathematical reasoning. In contrast, our method embeds the chain of self-correction, a strong reasoning mechanism, as an inherent capability in the LLMs through parameter fine-tuning. Furthermore, compared to existing SFT-based methods, our approach models mathematical reasoning as a multi-round procedure during the training stage. In addition, these SFT-based works \citep{yu2023teaching,an2023learning} perform the verification process in a straightforward manner, whereas our work conducts verification in a step-by-step manner by generating intermediate verification steps. Similar to the essence of the CoT approach, our model first verifies whether the generated code aligns with the problem description and then verifies whether the obtained result is consistent with the problem description. This step-by-step verification strategy significantly improves the verification process in self-correction. The excellent performance of our method on the challenging mathematical problem tasks, such as MATH and GSM8k, clearly demonstrates its effectiveness compared to the existing SFT-based methods \citep{yu2023teaching,an2023learning}. 

\section{Method}
\label{sec: method}

\begin{figure*}[t] \centering     
\includegraphics[width=\linewidth]{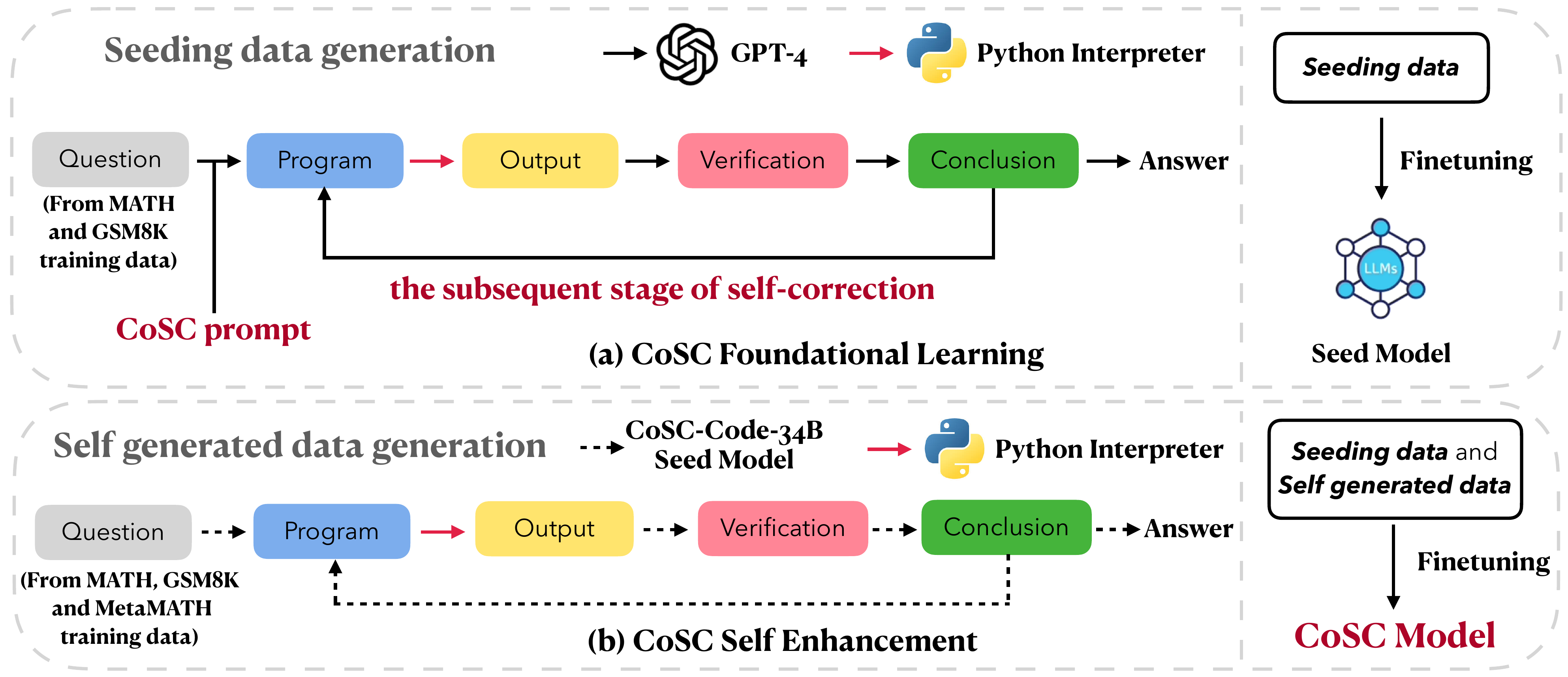} 
\vspace{-1.5em}
\caption{The training of Chain of Self-Correction (CoSC) consists of two phases. The first phase, \textbf{(a) CoSC Foundational Learning}, trains LLMs with seeding data generated from proprietary models, equipping them with a baseline proficiency in the CoSC methodology. In particular, we prompt GPT-4 with training questions from MATH \citep{hendrycks2021measuring} and GSM8K \citep{cobbe2021training} datasets to generate mathematical reasoning trajectories that adhere to the CoSC protocol. The second phase, \textbf{(b) CoSC Self Enhancement}, further adapts the seed model obtained from the previous phase with self-generated trajectories. These trajectories are produced by the seed model trained in the foundational phase, thereby enabling the generation of a substantial volume of data without the need for additional GPT-4 intervention. In both phases, we only retain trajectories whose answers match the ground-truth label.}
\vspace{-0.5em}
\label{fig:pipeline}
\end{figure*}

\subsection{Overview}

We propose a Chain of Self-Correction (CoSC) to address mathematical questions in a self-correction manner. 
Given a mathematical question denoted as $q$, the CoSC mechanism initiates the process by generating a program $p$ that comprises a Python function. The program $p$ is then executed using a Python interpreter to produce an output $o$. However, the program $p$ may contain bugs or incorrect intermediate code due to potential misinterpretations of the question $q$, leading to erroneous outputs. Unlike existing methods that either base their reasoning on incorrect intermediate content or halt reasoning upon encountering an error, the CoSC mechanism introduces a self-correction rationale during the reasoning process. After the generation of program $p$ and its output $o$, the CoSC model analyzes them and generates a verification $v$. This verification includes suggestions for improvements to ensure the consistency of the program $p$ and its output $o$ with the mathematical question $q$, respectively. Following this, the CoSC mechanism draws a conclusion $c$ based on these suggestions, which is used to either refine the program $p$ or generate the final answer. This process is repeated until a conclusive answer or a maximum limit of self-correction stages 
is reached, which can be summarized as $answer = povcpovc \cdots povc$. By employing this reasoning method, we can annotate public mathematical datasets and use the annotated multi-round, self-correction data to fine-tune LLMs.

\subsection{Training}

To enable the CoSC mechanism, we propose a two-phase finetuning method. (1) \textbf{CoSC foundational learning} with seeding data generated from proprietary models. (2) \textbf{CoSC self enhancement} with self-generated data using the seed model obtained in the first training phase. A summary for training our CoSC is shown in Fig. \ref{fig:pipeline}.

\subsubsection{CoSC Foundational Learning}

Existing mathematical reasoning datasets, such as chain-of-thoughts (CoT) \citep{wei2022chain} and program-of-thoughts (PoT) \citep{chen2022program}, primarily contain single-round annotations without multi-round, self-correction solutions for the reasoning process. This makes it challenging to fine-tune models to inherently possess self-correction capabilities. To address this issue, we utilize GPT-4 ($\mathcal{G}$) and a few-shot approach to annotate questions from publicly available mathematical datasets, including  MATH \citep{hendrycks2021measuring} and GSM8K \citep{cobbe2021training}, to generate seeding data with our Chain of Self-Correction (CoSC) mechanism. 
Detailed instructions and example cases of our CoSC can be found in Appendix \ref{app: Prompt for our CoSC} and Appendix \ref{app: Case of our CoSC on MATH and GSM8K Testing Sets}.


Given a mathematical question $q$, the few-shot prompt used to call GPT-4 is defined as $\wp$. The corresponding trajectory, denoted as $\tau$, is generated through the following process.
Firstly, we feed both the few-shot prompt $\wp$ and the question $q$  into GPT-4, which then generates a Python function code $p$ to solve the question $q$. It can be represented as:
\begin{equation}
\label{equ_a}
    \wp \oplus q \oplus \tau_{i-1} \to p_{i},
\end{equation}
where the symbol $\oplus$ represents concatenation and $\tau_i$ indicates the trajectory in the $i$ interaction round. 
After obtaining the code $p_i$, we execute it using a Python interpreter to acquire the runtime result $o_i$. 
Then, we enclose the result $o$ within ``\texttt{\char96}\texttt{\char96}\texttt{\char96}output'' and feed it, along with the previous few-shot prompt $\wp$, question $q$, the previous trajectory $\tau_i$, and generated code $p_i$ to the GPT-4. This produces a verification $v_i$ that analyzes the consistency between the question $q$ and the program code $p_i$, as well as between the question $q$ and the program output $o_i$.
Additionally, it produces a conclusion $c_i$ on whether to proceed with the next round of reasoning to perform adjustment, denoted as:
\begin{equation}
\label{equ_v}
    \wp\oplus q \oplus \tau_{i-1} \oplus p_{i}\oplus o_{i} \to v_{i} \oplus c_{i}.
\end{equation}
Depending on the conclusion $c_i$, it decides whether to terminate the reasoning process or proceed to the next round. If there is no inconsistency found during the verification step, it generates a natural language conclusion $c_i$ and encloses the answer to the problem with the special symbol ``boxed\{\}''. Otherwise, it should further revise the generated code $p_{i}$ in a new reasoning round by updating the trajectory $\tau_i$, denoted as:
\begin{equation}
\label{equ_tau}
    \tau_i = \tau_{i-1} \oplus p_{i} \oplus o_{i} \oplus v_{i} \oplus c_{i}.
\end{equation}

Based on the above steps, CoSC iteratively generates PoT code, program output, verification, and conclusion. At final, the reasoning process is stopped by checking whether the generated result contains an answer enclosed in ``boxed\{\}''. 

To ensure that the self-correction can concisely and clearly analyze the code and its runtime results, we design the self-correction in a step-by-step format as follows: 
\vspace{-0.5em}
\begin{itemize}
    \item The first step involves verifying whether the generated program code $p$ is consistent with the question $q$, such as checking the variables and their relationships.
    \item The second step involves verifying whether the code runtime results $o$ meet the requirements of the  question $q$, such as checking the reasonableness of numerical values.
\end{itemize}
\vspace{-0.5em}

Finally, for summary, CoSC will generate a conclusion $c$ to determine whether we should start the next round of reasoning or provide the final answer.


With CoSC, we use GPT-4 to annotate the  MATH and GSM8K training datasets. To explore diverse data, we apply nucleus sampling \citep{holtzman2019curious} during GPT-4 annotation. Each question is sampled in 3 times. However, for some complex questions, we are unable to obtain even a single correct solution. For these problems, we apply 10 more samplings and retained up to 4 correct data. Finally, we filtered out incorrect answers and constructed 37k pieces of data using GPT-4. 
The algorithm of the generation with our CoSC for each question from the training set is shown in Algorithm \ref{alg:interleave}.


Based on the 37k data constructed by GPT-4 as the seeding data, we apply CoSC foundational learning to train models. For a given question $q$, the response trajectory $\tau$ generated by GPT-4, denoted as $\tau=povcpovc \cdots povc$, the minimized negative log-likelihood loss used for training the model can be represented as:
\begin{equation}
arg\min_{\theta}\sum_{q,\tau}\sum_{i=1}^{n_q-1}-\text{log}\ \mathbb{P}_{\theta}(p_{i+1}o_{i+1}v_{i+1}c_{i+1}|q,p_i\cdots o_iv_ic_i),
\end{equation}
where $n_q$ is the iteration rounds of the question $q$ in our CoSC for the mathematical reasoning.


\begin{algorithm}[t]
\small
\caption{Inference Reasoning with our CoSC}
\label{alg:interleave}
    \textbf{Input:} question $q$, model $\mathcal{G}$, prompt $\wp$, external tools $\mathcal{E}$, stop condition \textit{Stop($\cdot$)}, iteration rounds $n_q$
    \begin{algorithmic}[1]

    \STATE $\tau_{0} \leftarrow \texttt{""}$ 
    
    \FOR{$i \leftarrow 1$ to $n_q$}
    
    \STATE $p_{i}\sim \mathbb{P}_\mathcal{G}(\cdot|\wp\oplus q\oplus \tau_{i-1})$ 
    
    \STATE $o_{i} \leftarrow \mathcal{E}(p_i)$ 
    
    \STATE $v_i \oplus c_i \sim \mathbb{P}_\mathcal{G}(\cdot|\wp\oplus q\oplus \tau_{i-1}\oplus p_i\oplus o_i)$ 
    
    \STATE $\tau_{i} \leftarrow \tau_{i-1}\oplus p_i\oplus o_{i}\oplus v_{i}\oplus c_{i}$ 
    
    \IF{\textit{Stop}} 
    \STATE  $\textbf{return}$ $\tau_{i}$
    \ENDIF
    
    \ENDFOR
    
    \STATE $\textbf{return}$ $\tau_{n}$
    
    \end{algorithmic}
\end{algorithm}



\subsubsection{CoSC Self Enhancement}

After the completion of the CoSC foundational learning, the seed model gains the ability to self-correct during inference and perform multi-round reasoning. Subsequently, we employ the seed model, after CoSC foundational learning, to apply dense solution sampling and dense question sampling, which enables the generation of more self-generated data with self-correction mechanisms.

\begin{table*}[ht]
\caption{Accuracy results (\%) on  MATH and GSM8K datasets. Vanilla models are tested with CoT.   ZS indicates the zero-shot inference without demonstrations. 
PAL refers to the Program-Aided Language model prompting \citep{gao2023pal}. The best results in each section are  in \textbf{bold} and the second-best results are \underline{underlined}. 
}
\vspace{-1em}
\label{tab:main}
\small
\centering
\setlength{\tabcolsep}{10pt}{
\begin{tabular}{lrll|ll|c}
\toprule
\textbf{Model} & \textbf{Base} &  \textbf{Size} & \textbf{ZS} & \ \ \textbf{MATH} & \textbf{GSM8K} & \textbf{AVG} \\ 
\midrule
\multicolumn{7}{c}{Proprietary Models} \\
\midrule
GPT-4o & - & - & \xmark & \ \ \textbf{76.6} \tiny{(0-shot CoT)} & 96.1 \tiny{(8-shot CoT)} & \textbf{86.4} \\
GPT-4V  & - & - & \xmark & \ \ 52.9 \tiny{(4-shot)} & 92.0 \tiny{(5-shot CoT)} & 72.5 \\
GPT-4 (PAL)  & - & - & \xmark & \ \  51.8 \tiny{(PAL)} & 94.2 \tiny{(PAL)} &  73.0 \\
GPT-4 & - & - & \xmark  & \ \  42.5 \tiny{(CoT)} &  92.0 \tiny{(5-shot CoT)} & 67.3 \\
ChatGPT (PAL)  & - & - & \xmark & \ \  {38.7} \tiny{(PAL)} & {78.6} \tiny{(PAL)} & 58.7 \\
ChatGPT & - & - & \xmark & \ \  35.5 \tiny{(CoT)} & {80.8} \tiny{(5-shot CoT)} & 58.2 \\
Claude-3.5 Sonnet & - & - & \xmark & \ \  71.1 \tiny{(0-shot CoT)} & \textbf{96.4} \tiny{(0-shot CoT)} & 83.8 \\
Claude-3 Opus & - & - & \xmark & \ \  60.1 \tiny{(0-shot CoT)} & 95.0 \tiny{(0-shot CoT)} & 77.6 \\
Gemini-1.5 Pro & - & - & \xmark & \ \  67.7 \tiny{(4-shot Minerva)} & 90.8 \tiny{(11-shot)} & 79.3  \\
Gemini-1.5 Flash & - & - & \xmark & \ \  54.9 \tiny{(4-shot Minerva)} & 86.2 \tiny{(11-shot)} & 70.6  \\
Gemini-1.0 Ultra & - & - & \xmark & \ \  53.2 \tiny{(4-shot Minerva)} & 88.9 \tiny{(11-shot)} &  71.1 \\
 \midrule
\multicolumn{7}{c}{Open-Source Models} \\
 \midrule
LLaMA-2 & LLaMA-2 & 7B &  \xmark & \ \ 4.1 \tiny{(CoT)} & 13.3 \tiny{(CoT)} & 8.7 \\
LLaMA-2 SFT & LLaMA-2 & 7B &  \cmark & \ \ 7.2 & 41.3 & 24.3 \\
LLaMA-2 RFT & LLaMA-2 & 7B & \cmark & \ \ - & 51.2 & - \\
CodeLLaMA (PAL) & CodeLLaMA & 7B &  \xmark & \ \ 16.6 \tiny{(PAL)} & 34.0 \tiny{(PAL)} & 25.3 \\
Platypus-2 \citep{lee2023platypus} & LLaMA-2 & 7B & \xmark & \ \ 5.4 \tiny{(CoT)} & 14.4 \tiny{(CoT)} & 9.9 \\
WizardMath \citep{luo2023wizardmath} & LLaMA-2 & 7B & \cmark & \ \ 10.7 & 54.9 & 32.8 \\
MetaMath \citep{yu2023metamath} & LLaMA-2 & 7B & \cmark & \ \ 19.8 & 66.5 & 43.2 \\
ToRA \citep{gou2023tora} & LLaMA-2 & 7B &  \cmark & \ \ 40.1 & 68.8 & 54.5 \\
CoSC (Ours) & LLaMA-2 & 7B &  \cmark & \ \ 42.7 & 70.5 & 56.6 \\
ToRA-Code \citep{gou2023tora} & CodeLLaMA & 7B  & \cmark & \ \ \underline{44.6} & \underline{72.6} & \underline{58.6} \\
CoSC-Code (Ours) & CodeLLaMA & 7B &  \cmark & \ \ \textbf{47.9} & \textbf{75.1} & \textbf{61.5 (+2.9)} \\
\midrule

LLaMA-2 & LLaMA-2 & 13B &  \xmark & \ \ 6.3 \tiny{(CoT)} & 24.3 \tiny{(CoT)} & 15.3 \\
LLaMA-2 SFT & LLaMA-2 & 13B &  \cmark & \ \ 9.2 &  51.1 & 30.2 \\
LLaMA-2 RFT & LLaMA-2 & 13B & \cmark & \ \  - & 55.3 &- \\
CodeLLaMA (PAL) & CodeLLaMA & 13B &\xmark & \ \ 19.9 \tiny{(PAL)} & 39.9 \tiny{(PAL)} & 29.9 \\
Platypus-2 \citep{lee2023platypus} & LLaMA-2 & 13B &  \xmark & \ \ 7.1 \tiny{(CoT)} & 23.7 \tiny{(CoT)} & 15.4 \\
WizardMath \citep{luo2023wizardmath} & LLaMA-2 & 13B &  \cmark & \ \  14.0 & 63.9 & 39.0 \\
MetaMath \citep{yu2023metamath} & LLaMA-2 & 13B & \cmark & \ \ 22.4 & 72.3 & 47.4 \\
ToRA \citep{gou2023tora} & LLaMA-2 & 13B & \cmark & \ \ 43.0 & 72.7 & 57.9 \\
CoSC (Ours) & LLaMA-2 & 13B &  \cmark & \ \ 45.3 & 73.9 & 59.6 \\
ToRA-Code \citep{gou2023tora} & CodeLLaMA & 13B &  \cmark & \ \ \underline{48.1} & \underline{75.8} & \underline{62.0} \\
CoSC-Code (Ours) & CodeLLaMA & 13B &  \cmark & \ \ \textbf{50.6} & \textbf{77.9} & \textbf{64.3 (+2.3)} \\

\midrule
CodeLLaMA (PAL)  & CodeLLaMA & 34B & \xmark & \ \ 23.9 \tiny{(PAL)} & 53.3 \tiny{(PAL)} & 38.6 \\
ToRA-Code \citep{gou2023tora} & CodeLLaMA & 34B & \cmark & \ \ \underline{50.8} & \underline{80.7} & \underline{65.8} \\
CoSC-Code (Ours) & CodeLLaMA & 34B & \cmark & \ \ \textbf{53.5} & \textbf{82.3} & \textbf{67.9 (+2.1)} \\

\bottomrule
\end{tabular}}
\vspace{-1em}
\end{table*}

\textbf{Dense solution sampling}. For mathematical questions, there are usually multiple solutions, but using GPT-4 for annotating them is expensive. 
Therefore, we use the seed model after CoSC foundational learning to resample the questions in the datasets multiple times, which further improves the model generalization ability in answering questions. 
Specifically, we use the CodeLLaMA-34B model which has the best performance after CoSC foundational learning to perform dense nucleus-sampling on 16k training data questions. 
Each question in the  MATH and GSM8K training datasets is sampled 64 times. 
We filter out the correct answers based on whether they match the ground-truth.

\textbf{Dense question sampling}. Similarly, there are multiple ways to ask a mathematical question. To improve the model generalization ability of question understanding, we use the data from MetaMath~\citep{yu2023metamath} to generalize the questions. 
Each question in MetaMATH is rewritten by simply rephrasing it, self-verifying the conditions in the question, adding if-then questions to reverse the question conditions, \textit{etc.}
Specifically, we use the CodeLLaMA-34B model with CoSC foundational learning to perform nucleus-sampling on the questions, sampling each question once, and filtering out data points based on whether they match the ground-truth.

In the end, we obtain a total of 339k data points, including 37k seeding data generated from GPT-4 and 302k generated from the CodeLLaMA-34B model with CoSC foundational learning. Then, we adopt them to train models from scratch to obtain our final CoSC model.

\subsection{Implementation Details}
\label{sec: Implementation Details}
By using 339k data points, including 37k seeding data and 302k self-generated data, we fine-tune the base models of LLaMA-2 \citep{touvron2023llama} and CodeLLaMA \citep{roziere2023code} to obtain our CoSC and CoSC-Code, respectively. They have different parameter sizes, such as 7B, 13B, and 34B. All models use full-scale fine-tuning. We use the AdamW optimizer with a learning rate of 2e-5 for all models, with a batch size set to 128, training for 1 epoch. To enable training, we use DeepSpeed ZeRO stage 3 \citep{rajbhandari2021zero} and Flash-Attention 2 \citep{dao2023flashattention} to optimize the model's memory usage. 
During inference, we set a maximum of 3 calls to the Python interpreter and a maximum token length of 2048.
The GPT-4 version for CoSC data generation is \texttt{gpt-4-0613}.
Our experiments train the models in 7B size with 8 NVIDIA A800 80GB GPUs and train the models in 13B and 34B with 16 NVIDIA A800 80GB GPUs.


\section{Experiments}
\label{sec: experiments}

\subsection{Evaluation Setup}
\label{sec: Datasets and Metric}
\textbf{Datasets}.
We evaluated models on the most widely used mathematical problem tasks, MATH \citep{hendrycks2021measuring} and GSM8K \citep{cobbe2021training}. 
The MATH dataset encompasses a total of 12,500 problems, partitioned into 7,500 for training and 5,000 for testing.
The GSM8K dataset contains 8,500 problems, with 7,500 for training and 1,000 for testing.
These datasets collectively encompass a broad range of mathematical questions, from basic arithmetic to competition level. More details of datasets are provided in Appendix \ref{app: Datasets Details}.

\textbf{Baselines}. For proprietary models, we present results from an array of SoTA LLMs, such as OpenAI's ChatGPT (gpt-3.5-turbo), OpenAI's GPT-4, OpenAI's GPT-4V, OpenAI's GPT-4o, Google's Gemini-1.0 Ultra, Google's Gemini-1.5 Flash, Google's Gemini-1.5 Pro, Anthropic's Claude-3 Opus, and Anthropic's Claude-3.5 Sonnet. By default, we report CoT prompting results, and include PAL (Program-Aided Language model) prompting \citep{gao2023pal} results for selected models. For open-source models, base models comprise LLaMA-2 \citep{touvron2023llama} and CodeLLaMA \citep{roziere2023code} with CoT and PAL prompting \citep{gao2023pal}. Supervised Fine-Tuning (SFT) employs CoT rationales from the original MATH and GSM8K dataset (15k samples) for fine-tuning. Rejection sampling Fine-Tuning (RFT) \citep{yuan2023scaling} leverages multiple models to generate diverse reasoning paths for fine-tuning. WizardMath \citep{luo2023wizardmath} augments data using ChatGPT, and conducts SFT and RLHF. Platypus-2 \citep{lee2023platypus} is fine-tuned with Open-Platypus reasoning datasets. ToRA \citep{gou2023tora} uses GPT-4 to generate tool-used trajectories and finetunes on it. The prompting methods for each evaluation are shown in Appendix \ref{app: Prompting Methods for Each Evaluation}.

\textbf{Metric}.
We report accuracies of predicted answers. For numerical values, we perform rounding, while for expressions, we employ the Python library \textit{sympy} for parsing. 

\subsection{Main Results}

In our experiments, we observe several findings that demonstrate the effectiveness of our proposed CoSC model on mathematical datasets. We show the experimental results in Table \ref{tab:main}.
First and foremost, our proposed CoSC consistently outperforms previous state-of-the-art open-source LLMs across all scales. Specifically, our CoSC-Code can achieve an average improvement of 2.9\%, 2.3\%, and 2.1\%, on 7B, 13B, and 34B size, respectively.

Moreover, to further emphasize the superiority of our CoSC, we conduct a comprehensive comparison against multiple proprietary models. The results reveal that our CoSC-Code-34B can outperform all the advanced proprietary LLMs, such as ChatGPT and GPT-4, as well as most advanced proprietary multi-modal LLMs, such as GPT-4V, Gemini-1.0 Pro, and Gemini-1.0 Ultra on MATH dataset.  
It is important to note that, unlike these proprietary models, our CoSC performs the inference in a zero-shot manner without demonstrations.

\subsection{Ablation Study}
In our ablation study, we conduct experiments on MATH dataset with the largest number of 5,000 test samples with broad spectrum of subjects and difficulty levels. For the consideration of computational efficiency, we choose the smaller models of CodeLLaMA with size 7B and 13B as the base models.

\subsubsection{Effect of Each Training Phase}
Our CoSC mechanism comprises two  training phases: (1) CoSC foundational learning and (2) CoSC self-enhancement. In order to assess the individual contributions of each training phase, we evaluate multiple combinations of the proposed training phases, as shown in Table \ref{tab:ablation}. 

The experiments conducted on CodeLLaMA serve as the baseline for our study. Initially, we employ GPT-4 to generate 37k seeding data for CoSC foundational learning, which equips LLMs with initial self-correction capabilities. As a result, we observe a notable improvement in performance on the MATH dataset, with accuracy increasing from 16.6\% to 42.3\% for the 7B size and from 19.9\% to 47.0\% for the 13B size.

To further enhance the self-correction performance, we utilize the seed model obtained in the first phase to generate additional data without relying on GPT-4. This approach leads to further 
improvements in performance on the MATH dataset from 42.3\% to 47.9\% and 47.0\% to 50.6\%, for 7B and 13B respectively.

It is exciting to see that the originally weak LLMs, such as the CodeLLaMA 7B base model and the 13B base model, can significantly improve in mathematical reasoning by using the proposed CoSC method, as shown in Table \ref{tab:ablation}. This demonstrates the effectiveness of the CoSC method, which is able to embed self-correction as an inherent capability in LLMs, leading to significant improvements in mathematical reasoning performance.

\subsubsection{Effect of Multi-Round Reasoning in the Proposed CoSC mechanism}
Our CoSC mechanism integrates a series of self-correction stages to progressively verify and refine output of LLMs. In particular, the conclusion step in our CoSC relies on clues from the verification step to determine whether to proceed to the next round of reasoning or directly provide the final answer. Such iterative multi-round mechanism enables LLMs to self-correct their outputs and improve accuracy. This experiment quantitatively investigates effect of this multi-round mechanism. As shown in Table \ref{tab:multi step reasoning}, our CoSC can effectively generate more rounds of mathematical reasoning during inference, confirming efficacy of our CoSC mechanism in enhancing the reasoning process.

In addition, we conduct a comparison between single-round reasoning and multi-round reasoning using our CoSC mechanism on the test samples from the MATH dataset. The single-round results are obtained by extracting the answer solely from the output of the first round, without any self-correction enabled. To illustrate the impact of multi-round reasoning, we present the comparative results in Table \ref{tab:multi step performance}. The results clearly demonstrate the effectiveness of multi-round reasoning in rectifying errors and improving overall accuracy. With the 7B model, we observe an improvement in accuracy from 40.2\% to 47.9\%, while with the 13B model, accuracy increases from 42.4\% to 50.6\%. These findings highlight the significant benefits of employing multi-round reasoning within our CoSC mechanism. 
More ablation studies are shown in Appendix \ref{app: More Ablation Studies}.

\begin{table}[t]
\begin{minipage}{\linewidth}
\caption{Accuracy results (\%) on the MATH   dataset for the CoSC foundational learning and  CoSC self enhancement. The results show the superiority and necessity of both training phases.
}
\vspace{-1em}
\label{tab:ablation}
\small
\centering
\setlength{\tabcolsep}{7pt}{
\begin{tabular}{cc|cc}
\toprule
\multicolumn{2}{c|}{Chain of Self-Correction (CoSC)} & \multicolumn{1}{c}{\multirow{2}{*}{7B}}  & \multicolumn{1}{c}{\multirow{2}{*}{13B}} \\
Foundational Learning &  Self Enhancement & &   \\
\cmidrule{1-4}
\xmark & \xmark & 16.6 & 19.9  \\
\cmark & \xmark & 42.3 & 47.0  \\ 
\cmark & \cmark & 47.9 & 50.6  \\ 
\bottomrule
\end{tabular}}
\end{minipage}
\vspace{0.5em}

\begin{minipage}{\linewidth}
\caption{The distribution of reasoning rounds on MATH test dataset across three LLMs
in 7B and 13B size. Our CoSC-Code can generate more rounds compared to other models in mathematical reasoning. \#Round indicates the number of reasoning rounds during inference.
}
\vspace{-1em}
\label{tab:multi step reasoning}
\small
\centering
\setlength{\tabcolsep}{3.5pt}{
\begin{tabular}{lc|ccc}
\toprule
Models & Size & \#Round=1 & \#Round=2 & \#Round=3 \\
\cmidrule{1-5}
\multicolumn{1}{l}{\multirow{2}{*}{CodeLLaMA}}  & 7B & 100\% & 0 & 0 \\
 & 13B & 100\% & 0 & 0 \\ \midrule
\multicolumn{1}{l}{\multirow{2}{*}{ToRA-Code}}  & 7B & 100\% & 0 & 0 \\ 
 & 13B & 100\% & 0 & 0 \\  \midrule
\multicolumn{1}{l}{\multirow{2}{*}{CoSC-Code (Ours)}}  & 7B & 78.3\% & 12.7\% & 9.0\%    \\ 
 & 13B & 79.3\% & 13.1\% & 7.6\% \\
\bottomrule
\end{tabular}}
\end{minipage}
\vspace{0.5em}

\begin{minipage}{\linewidth}
\caption{Accuracy results (\%)  for the test samples on the MATH dataset of our CoSC-Code with single-round reasoning and multi-round reasoning during the inference stage. 
}
\vspace{-1em}
\label{tab:multi step performance}
\small
\centering
\setlength{\tabcolsep}{5.8pt}{
\begin{tabular}{l|cccc}
\toprule
Models & 7B & 13B  \\
\cmidrule{1-3}
 CoSC-Code with single-round reasoning & 40.2 & 42.4 \\ 
 CoSC-Code with multi-round reasoning (Ours) & 47.9 & 50.6 \\ 
\bottomrule
\end{tabular}}
\end{minipage}
\vspace{-1em}
\end{table}

\section{Conclusion}
\label{sec: conclusion}



In conclusion, our Chain of Self-Correction (CoSC) mechanism equips LLMs with the ability to autonomously validate and refine their outputs. This mechanism facilitates a sequence of self-correction stages that progressively refine the reasoning process, leading to enhanced accuracy in mathematical reasoning. 
Through extensive experiments, we have demonstrated the remarkable performance improvement that CoSC brings to mathematical datasets.
We believe that our CoSC mechanism can provide valuable insights for future research 
on LLMs across various domains.

\section*{Impact Statement}

We experiment on two mathematical datasets, including GSM8K and MATH, both of which use MIT License code. The prompts used in these experiments are listed in Appendix \ref{app: Prompt for our CoSC}, and we want to emphasize that none of the prompts contain any words that discriminate against any individual or group. Furthermore, prompts would not negatively impact anyone’s safety in this work.





\nocite{langley00}

\bibliography{ms}
\bibliographystyle{icml_arxiv}

\newpage
\appendix
\onecolumn
\section{Prompt for our CoSC}
We present specific instructions and example few-shot prompts of our CoSC for querying GPT-4 to generate the seeding data.
\label{app: Prompt for our CoSC}

\lstset{
    language=Python,
    basicstyle=\ttfamily\small,
    keywordstyle=\color{blue},
    stringstyle=\color{red},
    commentstyle=\color{green},
    showstringspaces=false,
    breaklines=true  
}

\subsection{Prompt for our CoSC on MATH}
The prompt for CoSC on MATH is as follows:

\begin{lstlisting}
Integrate step-by-step reasoning and Python code to solve math problems using the following guidelines:

 - Analyze the question and write functions to solve the problem; the function should not take any arguments.

 - Present the final results in LaTeX using a "\boxed{}" without any units.

 - Utilize the "pi" symbol and "Rational" from Sympy for $\pi$ and fractions, and simplify all fractions and square roots without converting them to decimal values.

 - In "verification", verify the consistency between the "Question" and the "python" code, as well as between the "Question" and "output" in "Solution"  step-by-step.

 - According to the "verification", rewrite the code if neccessary.
    
Here are some examples you may refer to:

---

Question: Find the matrix that corresponds to projecting onto the vector $\begin{pmatrix} 2 \\ -3 \end{pmatrix}.$

Solution:

```python
from sympy import Matrix, simplify

def projection_matrix():
    v = Matrix([2, -3])

    # Calculate the projection matrix
    P = (v * v.T) / (v.T * v)[0]

    # Simplify the elements
    P = P.applyfunc(simplify)

    return P

answer = projection_matrix()
print(answer)
```
\end{lstlisting}
\begin{lstlisting}[language=TeX]
```output
Matrix([[4/13, -6/13], [-6/13, 9/13]])
```

```verification
Step 1. Verify the consistency between the question and the code. The "python" code defines a function binary_sum_diff() that calculates the projection matrix for the given vector. The vector is defined as `v = Matrix([2, -3])`, and then use standard formula `(v * v.T) / (v.T * v)[0]` for calculating the projection matrix onto a vector, finally use `applyfunc(simplify)` to simplify the elements of the matrix.
Step 2: Verify the consistency between the question and the output. The answer matrix of 2 dimension vector should be a 2x2 matrix, the "output" of "python" code is `[[4/13, -6/13], [-6/13, 9/13]]`, which is reasonable.
```

```conclusion
Therefore, the "python" code and "output" are consistent with "Question". The matrix is $\boxed{\begin{pmatrix} \frac{4}{13} & -\frac{6}{13} \ -\frac{6}{13} & \frac{9}{13} \end{pmatrix}}.$
```
\end{lstlisting}

\begin{lstlisting}
---

Question: Compute $1011_2 + 101_2 - 1100_2 + 1101_2$.  Express your answer in base 2.

Solution:

```python
def binary_sum_diff():
    # Convert binary to decimal
    num1 = int("1011", 2)
    num2 = int("101", 2)
    num3 = int("1100", 2)
    num4 = int("1101", 2)

    # Perform the operations
    result = num1 + num2 - num3 + num4

    # Convert the result back to binary
    result_binary = format(result, "b")

    return result_binary

answer = binary_sum_diff()
print(answer)
```
\end{lstlisting}
\begin{lstlisting}[language=TeX]
```output
10001
```

```verification
Step 1. Verify the consistency between the question and the code. The "python" code defines a function binary_sum_diff() that convert the binary numbers to decimal, performing the operations, and then converting the result back to binary. It assigns the decimal form of values $1011, 101, 1100, 1101$ to $num1, num2, num3, num4$ respectively, and then perform the operation $num1 + num2 - num3 + num4, finally convert the result back to binary.
Step 2: Verify the consistency between the question and the output. The "output" of "python" code is $10001$, which is reasonable.
```

```conclusion
Therefore, the "python" code and "output" are consistent with "Question". The answer expressed in base 2 is $\boxed{10001_2}$.
```
\end{lstlisting}

\begin{lstlisting}
---

Question: Convert the point $(0, -3 \sqrt{3}, 3)$ in rectangular coordinates to spherical coordinates.  Enter your answer in the form $(\rho,\theta,\phi),$ where $\rho > 0,$ $0 \le \theta < 2 \pi,$ and $0 \le \phi \le \pi.$

Solution:

```python
from sympy import sqrt, atan2, acos, pi

def rectangular_to_spherical():
    x, y, z = 0, -3*sqrt(3), 3
    rho = sqrt(x**2 + y**2 + z**2)
    theta = atan2(y, x)
    phi = acos(z/rho)
    return rho, theta, phi

answer = rectangular_to_spherical()
print(answer)
```
\end{lstlisting}
\begin{lstlisting}[language=TeX]
```output
(6, -pi/2, pi/3)
```

```verification
Step 1. Verify the consistency between the question and the code. The "python" code defines a function rectangular_to_spherical() that convert rectangular coordinates to spherical coordinates. It assigns the values $0, -3 \sqrt{3}, 3$ to $x, y, z$ respectively, and then use the following formulas $\rho = \sqrt{x^2 + y^2 + z^2}$, $\theta = \arctan{\frac{y}{x}}$ 3, $\phi = \arccos{\frac{z}{\rho}}$ to calculates $rho, theta, phi$.
Step 2: Verify the consistency between the question and the output. The required form of the "output" is $(\rho, \theta, \phi)$, where $\rho > 0$, $0 \le \theta < 2\pi$, and $0 \le \phi \le \pi$. The "output" of "python" code is $(6, -\pi/2, \pi/3)$, which is not reasonable.
```

```conclusion
Therefore, the "python" code and "output" are not consistent with "Question". Let's rewrite the "python" code based on the "verification":
```
\end{lstlisting}

\begin{lstlisting}
```python
from sympy import sqrt, atan2, acos, pi

def rectangular_to_spherical():
    x, y, z = 0, -3*sqrt(3), 3
    rho = sqrt(x**2 + y**2 + z**2)
    theta = atan2(y, x)
    phi = acos(z/rho)
    theta = (theta + 2 * pi) % (2 * pi)
    phi = (phi + pi) % pi
    return rho, theta, phi

answer = rectangular_to_spherical()
print(answer)
```
\end{lstlisting}
\begin{lstlisting}[language=TeX]
```output
(6, 3*pi/2, pi/3)
```

```verification
Step 1. Verify the consistency between the question and the code. The "python" code was rewritten, `theta = (theta + 2 * pi) % (2 * pi)` and `phi = (phi + pi) % pi` were added to meet requirement that $0 \le \theta < 2\pi$, and $0 \le \phi < \pi$.
Step 2: Verify the consistency between the question and the output. The required form of the "output" is $(\rho, \theta, \phi)$, where $\rho > 0$, $0 \le \theta < 2\pi$, and $0 \le \phi \le \pi$. The "output" of "python" code is $(6, 3*\pi/2, \pi/3)$, which is reasonable.
```

```conclusion
Therefore, the "python" code and "output" are consistent with "Question". Hence, the spherical coordinates for the point $(0, -3\sqrt{3}, 3)$ are $\boxed{(6, \frac{3\pi}{2}, \frac{\pi}{3})}$.
```
\end{lstlisting}

\begin{lstlisting}
---

Question: Solve the inequality
\[\frac{x^2 - 25}{x + 5} < 0.\]

Solution:

```python
from sympy import symbols, simplify
from sympy.solvers.inequalities import solve_univariate_inequality
from sympy.core.relational import LessThan

def solve_inequality():
    x = symbols('x')
    expression = (x**2 - 25) / (x + 5)
    inequality = LessThan(expression, 0)
    solution = solve_univariate_inequality(inequality, x, relational=False)
    simplified_solution = simplify(solution)
    
    return simplified_solution

answer = solve_inequality()
print(answer)
```
\end{lstlisting}
\begin{lstlisting}[language=TeX]
```output
Union(Interval.open(-oo, -5), Interval.Lopen(-5, 5))
```

```verification
Step 1. Verify the consistency between the question and the code. The "python" code defines a function solve_inequality() that that solves the inequality \[\frac{x^2 - 25}{x + 5} < 0.\]. It defines a symbolic variable `x` and the expression is defined as `\[\frac{x^2 - 25}{x + 5}\]`, then use function `LessThan` to define the inequality and function `solve_univariate_inequality` to solve the inequality for `x`, finally use `simplify` to simplify the result.
Step 2: Verify the consistency between the question and the output. The required answer `x` should meet the condition: $x + $5 != $0, the "output" of "python" code is `(-oo, -5)` and `(-5, 5)`, which is reasonable.
```

```conclusion
Therefore, the "python" code and "output" are consistent with "Question". The answer is $\boxed{(-\infty,-5)\cup(-5,5)}$
```
\end{lstlisting}

\begin{lstlisting}
---

Question: In the figure, triangles $ABC$ and $BCD$ are equilateral triangles. What is the value of $AD \div BC$ when expressed in simplest radical form?

\begin{asy}
draw((0,0)--(5,8.7)--(10,0)--cycle);
draw((10,0)--(15,8.7)--(5,8.7));
label("$A$",(0,0),SW);
label("$B$",(5,8.7),N);
label("$C$",(10,0),SE);
label("$D$",(15,8.7),NE);
\end{asy}


Solution:

```python
from sympy import Rational, sqrt, simplify

def ad_divided_by_bc():

    x = 1  # Side length of equilateral triangles

    ad_squared = 2 * x**2 - 2 * x**2 * cos(2 * pi / 3)  # Using the law of cosines
    ad = sqrt(ad_squared)

    bc = x # BC is the side length of the equilateral triangles

    simplified_ratio = simplify(ad / bc)

    return simplified_ratio

answer = ad_divided_by_bc()
print(answer)
```
\end{lstlisting}

\begin{lstlisting}[language=TeX]
```output
sqrt(3)
```

```verification
Step 1. Verify the consistency between the question and the code. The "python" code defines a function ad_divided_by_bc() that calculates the ratio of the lengths of $AD$ and $BC$ in the given figure. The side length of the equilateral triangles `x` is set to 1, because $ABC$ and $BCD$ are equilateral triangles, the angle between $AB$ and $AD$ should be 2 * \pi / 3, then we use the law of cosines, the square length of $AD$ is thus `2 * x**2 - 2 * x**2 * cos(2 * pi / 3)`, and the length of $AD$ is the square root of that, finally use `simplify` to simplify the 'ad / bc'.
Step 2: Verify the consistency between the question and the output. The required answer $AD \div BC$ should be positive, the "output" of "python" code is $\sqrt{3}$, which is reasonable.
```

```conclusion
Therefore, the "python" code and "output" are consistent with "Question". The value of $AD \div BC$ is $\boxed{\sqrt{3}}$.
```

---
\end{lstlisting}

\subsection{Prompt for our CoSC on GSM8K}
\label{app:A1}
The prompt for CoSC on GSM8K is as follows:

\begin{lstlisting}
Integrate step-by-step reasoning and Python code to solve math problems using the following guidelines:

 - Analyze the question and write functions to solve the problem; the function should not take any arguments.

 - Present the final results in LaTeX using a "\boxed{}" without any units.

 - Utilize the "pi" symbol and "Rational" from Sympy for $\pi$ and fractions, and simplify all fractions and square roots without converting them to decimal values.

 - In "verification", verify the consistency between the "Question" and the "python" code, as well as between the "Question" and "output" in "Solution"  step-by-step.

 - According to the "verification", rewrite the code if neccessary.
    
Here are some examples you may refer to:

---

Question: Olivia has \$23. She bought five bagels for \$3 each. How much money does she have left?

Solution:

```python
from sympy import simplify

def money_left():
    money_initial = 23
    bagels = 5
    bagel_cost = 3
    money_spent = bagels * bagel_cost
    remaining_money = simplify(money_initial - money_spent)
    return remaining_money
 
answer = money_left()
print(answer)
```
\end{lstlisting}

\begin{lstlisting}[language=TeX]
```output
8
```

```verification
Step 1. Verify the consistency between the question and the code. The "python" code defines a function "money_left()" that calculates the remaining money after Olivia buys the bagels. The initial money is set to 23, the number of bagels is set to 5, and the cost of each bagel is set to 3. The total cost of the bagels is calculated as bagels * bagel_cost, which is $5 * $3 = $15. The remaining money is calculated as money_initial - money_spent.
Step 2: Verify the consistency between the question and the output. The left money should be a positive number or zero, and the "output" of "python" code is $8 >= $0, which is reasonable.
```

```conclusion
Therefore, the "python" code and "output" are consistent with "Question". Olivia has $\boxed{8}$ dollars left.
```

---

Question: Michael had 58 golf balls. On tuesday, he lost 23 golf balls. On wednesday, he lost 2 more. How many golf balls did he have at the end of wednesday?

Solution:

```python
from sympy import simplify

def remaining_golf_balls():
    golf_balls_initial = 58
    golf_balls_lost_tuesday = 23
    golf_balls_lost_wednesday = 2
    golf_balls_left = golf_balls_initial - golf_balls_lost_tuesday - golf_balls_lost_wednesday
    remaining_golf_balls = simplify(golf_balls_left)
    return remaining_golf_balls

answer = remaining_golf_balls() 
print(answer)
```
\end{lstlisting}
\begin{lstlisting}[language=TeX]
```output
33
```

```verification
Step 1. Verify the consistency between the question and the code. The "python" code defines a function remaining_golf_balls() that calculates the remaining golf balls. The initial number of golf balls is set to 58, the lost on Tuesday is set to 23 and the lost on Wednesday is set to 2. The remaining golf balls is calculated as golf_balls_initial - lost_on_Tuesday - lost_on_Wednesday.
Step 2: Verify the consistency between the question and the output. The left number of golf balls should be a positive number or zero, and the "output" of "python" code is $33 >= $0, which is reasonable.
```

```conclusion
Therefore, the "python" code and "output" are consistent with "Question". Michael had $\boxed{33}$ golf balls at the end of Wednesday.
```

---

Question: There were nine computers in the server room. Five more computers were installed each day, from monday to thursday. How many computers are now in the server room?

Solution:

```python
from sympy import simplify

def total_computers():
    computers_initial = 9
    computers_per_day = 5
    num_days = 3
    computers_added = computers_per_day * num_days
    computers_total = simplify(computers_initial + computers_added)
    return computers_total

answer = total_computers()
print(answer)
```
\end{lstlisting}
\begin{lstlisting}[language=TeX]
```output
24
```

```verification
Step 1. Verify the consistency between the question and the code. The "python" code defines a function total_computers() that calculates the number of computers now. The initial number of computers is set to 9, the installed computer each day is set to 5, the number of days is set to 3, the number of computer now in the server room is calculated as computers_per_day * num_days + computers_initial, however there are 4 days from monday to thursday.
Step 2: Verify the consistency between the question and the output. The number of computers should be a positive number or zero, and the "output" of "python" code is $24 >= $0, which is reasonable.
```

```conclusion
Therefore, the "python" code is not consistent with "Question". Let's rewrite the "python" code based on the "verification":
```
\end{lstlisting}
\begin{lstlisting}
```python
from sympy import simplify

def total_computers():
    computers_initial = 9
    computers_per_day = 5
    num_days = 4 # from monday to thursday
    computers_added = computers_per_day * num_days
    computers_total = simplify(computers_initial + computers_added)
    return computers_total

answer = total_computers()
print(answer)
```
\end{lstlisting}
\begin{lstlisting}[language=TeX]
```output
29
```

```verification
Step 1. Verify the consistency between the question and the code. The "python" code was rewritten, `num_days` is set to 4, which consistent with the `from monday to thursday` in "Question".
Step 2: Verify the consistency between the question and the output. The number of computers should be a positive number or zero, and the "output" of "python" code is $29 >= $0, which is reasonable.
```

```conclusion
Therefore, the "python" code are consistent with "Question". There're $\boxed{29}$ computers in the server room.

---
```
\end{lstlisting}

\section{Case of our CoSC on MATH and GSM8K Testing Sets}
\label{app: Case of our CoSC on MATH and GSM8K Testing Sets}
We would like to show examples how our CoSC model generate a corresponding answer when a question is queried.
\subsection{Case on MATH testing set}
The case for CoSC on MATH testing set is as follows, where the question is input into our CoSC model and the solution represents the corresponding answer generated from our CoSC model.
\begin{lstlisting}
Question: Find the sum of all values of x such that \abs{x-1}=7.

Solution:

```python
from sympy import symbols, solve, Abs

def solve_absolute_equation():
    x = symbols('x')
    equation = Abs(x - 1) - 7
    solutions = solve(equation, x)
    sum_solutions = sum(solutions)
    return sum_solutions

answer = solve_absolute_equation()
print(answer)
```
\end{lstlisting}
\begin{lstlisting}
```output
Program output: NotImplementedError: solving Abs(x - 1) when  the argument is not real or imaginary.
```

```verification
Step 1. Verify the consistency between the question and the code. The "python" code defines a function solve_absolute_equation() that solves the absolute equation $|x-1| = 7$. It defines a symbolic variable `x` and the equation is defined as `Abs(x - 1) - 7`, then use function `solve` to solve the equation for `x`, finally use `sum` to get the sum of all solutions.
Step 2: Verify the consistency between the question and the output. The "output" of "python" code is `NotImplementedError: solving Abs(x - 1) when the argument is not real or imaginary.`, which is not reasonable.
```

```conclusion
Therefore, the "python" code and "output" are  not consistent with "Question". Let's rewrite the "python" code based on the "Verification".
```
\end{lstlisting}
\begin{lstlisting}
```python
from sympy import symbols, solve

def solve_absolute_equation():
    x = symbols('x')
    equation1 = x - 1 - 7
    equation2 = 1 - x - 7
    solutions1 = solve(equation1, x)
    solutions2 = solve(equation2, x)
    sum_solutions = sum(solutions1) + sum(solutions2)
    return sum_solutions

answer = solve_absolute_equation()
print(answer)
```
\end{lstlisting}
\begin{lstlisting}
```output
2
```

```verification
Step 1. Verify the consistency between the question and the code. The "python" code was rewritten, `equation1 = x - 1 - 7` and `equation2 = 1 - x - 7` were added to solve the two possible equations separately, and then get the sum of all solutions.
Step 2: Verify the consistency between the question and the output. The required answer is the sum of all solutions of the equation $|x-1| = 7$, the "output" of "python" code is $2$, which is reasonable.
```

```conclusion
Therefore, the "python" code and "output" are consistent with "Question". The sum of all values of $x$ such that $|x-1| = 7$ is $\boxed{2}$.
```
\end{lstlisting}

\subsection{Case on GSM8K testing set}
The case for CoSC on GSM8K testing set is as follows, where the question is input into our CoSC model and the solution represents the corresponding answer generated from our CoSC model.
\begin{lstlisting}
Question: A shop sells school supplies. One notebook is sold at \$1.50 each, a pen at \$0.25 each, a calculator at \$12 each, and a geometry set at \$10. Daniel is an engineering student, and he wants to buy five notebooks, two pens, one calculator, and one geometry set. The shop gives a 10\% discount on all the purchased items. How much does Daniel have to spend on all the items he wants to buy? (\textcolor{green}{Ground truth is 27})

Solution: 

```python
from sympy import simplify, Rational

def total_cost():
    notebook_cost = 1.5
    pen_cost = 0.25
    calculator_cost = 12
    geometry_set_cost = 10
    discount = Rational(10, 100)
    notebooks = 5
    pens = 2
    calculators = 1
    geometry_sets = 1
    total_cost_before_discount = notebook_cost * notebooks + pen_cost * pens + calculator_cost * calculators + geometry_set_cost * geometry_sets
    total_discount = total_cost_before_discount * discount
    total_cost_after_discount = simplify(total_cost_before_discount * (1 - discount))
    return total_cost_after_discount

answer = total_cost()
print(answer)
```
\end{lstlisting}
\begin{lstlisting}[language=TeX]
```output
27.0000000000000
```

```verification
Step 1. Verify the consistency between the question and the code. The "python" code defines a function total_cost() that calculates the total cost after discount. The cost of each item and the discount are set to their respective values. The total cost before discount is calculated as the sum of the cost of each item multiplied by their quantities. The total discount is calculated as total_cost_before_discount * discount. The total cost after discount is calculated as total_cost_before_discount - total_cost_after_discount.
Step 2: Verify the consistency between the question and the output. The total cost after discount should be a positive number or zero, and the "output" of "python" code is $27 >= $0, which is reasonable.
```

```conclusion
Therefore, the "python" code is consistent with "Question". Daniel has to spend $\boxed{27}$ dollars on all the items he wants to buy.
```
\end{lstlisting}

\section{Evaluation Setup Details}

\subsection{Datasets Details}
\label{app: Datasets Details}

\textbf{MATH} \citep{hendrycks2021measuring}. This dataset consists of competition level mathematics problems. It encompasses a total of 12,500 problems, partitioned into 7,500 for training and 5,000 for testing. Each problem is accompanied by a step-by-step solution and concludes with a distinct final answer, which is formatted for straightforward verification of the model generated solutions. Notably, the MATH dataset spans a broad spectrum of subjects and difficulty levels, including seven categories: Prealgebra, Algebra, Number Theory, Counting and Probability, Geometry, Intermediate Algebra, and Precalculus. 

\textbf{GSM8K} \citep{cobbe2021training}. Comprising a diverse collection of grade school mathematical word problems, GSM8K is recognized for its high quality. While it is generally considered less challenging than the MATH dataset, it similarly provides step-level solutions with basic arithmetic operations (addition, subtraction, multiplication, division). The GSM8K dataset contains 8,500 problems, with 7,500 for training and 1,000 for testing.

\subsection{Prompting Methods for Each Evaluation}
\label{app: Prompting Methods for Each Evaluation}

Table \ref{tab:main} includes an identifier ``ZS'', which denotes whether the LLMs are evaluated in a zero-shot inference setting without demonstrations. To clarify further, we summarize below the prompting methods employed for each evaluation.

Proprietary Models:

\begin{itemize}
    \item GPT-4o~\citep{openai2024gpt4o}: Zero-shot CoT prompting for MATH; 8-shot CoT prompting for GSM8K.
    \item GPT-4V~\citep{openai2023gpt4v}: 4-shot prompting for MATH; 5-shot CoT prompting for GSM8K.
    \item GPT-4 and ChatGPT~\citep{openai2023gpt4}: CoT prompting for MATH; 5-shot CoT prompting for GSM8K.
    \item Gemini family~\citep{team2024gemini}: 4-shot Minerva prompting for MATH; 11-shot prompting for GSM8K.
    \item Claude family~\citep{anthropic2024Claude}: Zero-shot CoT prompting for both datasets.
    \item PaLM-2~\citep{anil2023palm}: 4-shot CoT prompting for MATH; 8-shot CoT prompting for GSM8K.
\end{itemize}

Open-Source Models:

\begin{itemize}
    \item LLaMA-2~\citep{touvron2023llama} and Platypus-2~\citep{lee2023platypus}: CoT prompting for both datasets.
    \item CodeLLaMA~\citep{roziere2023code}: Program-Aided Language (PAL)~\citep{gao2023pal}  prompting for both datasets.
    \item LLaMA-2 SFT~\citep{yuan2023scaling}, LLaMA-2 RFT~\citep{yuan2023scaling}, WizardMath~\citep{luo2023wizardmath}, MetaMath~\citep{yu2023metamath}, ToRA~\citep{gou2023tora}, and our CoSC method: Fully zero-shot, requiring no demonstrations.
\end{itemize}

\section{More Ablation Studies}
\label{app: More Ablation Studies}





\subsection{Accuracy of the Verification Module and Error Reduction of the Correction Module}

Our CoSC framework comprises two main components: verification and correction. The verification component identifies potential erroneous reasoning, while the correction component generates improved reasoning to address the issues identified in the verification step. To evaluate the effectiveness of these components, we conducted ablation studies. We provide a detailed analysis of the benefits derived from both modules, reporting the accuracy of the verification module and the error reduction achieved by the correction module. The accuracy of the verification module reflects how precisely it identifies errors, whereas the error reduction of the correction module measures the rate at which errors are corrected from one round to the next. Specifically, it examines the proportion of solutions deemed incorrect in the i-th round that are successfully corrected in the (i + 1)-th round.

The results of CoSC-Code in 7B and 13B on  MATH dataset are shown in Table \ref{tab:the accuracy of the verification module and the error reduction achieved by the correction module}. It can be observed that the verification of our CoSC-Code can iteratively refine its outputs with high accuracy, about 70\%. Furthermore, for more difficult questions that require self-correction across multiple rounds, our CoSC-Code is still capable of successfully reducing errors by over 25\%. This confirms the effectiveness of both the verification module and the correction module in our CoSC method.

\subsection{Context Lengths for Different Questions}

We explore the effect of the context lengths when addressing different questions over multiple rounds of self-corrections. We have calculated the percentage distribution of context lengths for CoSC-Code in 7B and 13B on MATH dataset, as shown in Table \ref{tab:the effect of the context lengths}. These statistics indicate that all context lengths fall within the 4K range, which is well-suited for modern LLMs.


\begin{table}[t]
\begin{minipage}{\linewidth}
\caption{The accuracy of the verification module and the error reduction rate achieved by the correction module of CoSC-Code in 7B and 13B on MATH dataset.
}
\vspace{-1em}
\label{tab:the accuracy of the verification module and the error reduction achieved by the correction module}
\small
\centering
\setlength{\tabcolsep}{10pt}{
\begin{tabular}{l|cccc}
\toprule
Models & accuracy of the verification module & error reduction of the correction module  \\
\cmidrule{1-3}
CoSC-Code-7B & 68.5\% &  25.74\%  \\
CoSC-Code-13B &  70.4\% & 26.25\% \\
\bottomrule
\end{tabular}}
\end{minipage}
\vspace{1em}

\begin{minipage}{\linewidth}
\caption{The context lengths when addressing different questions over multiple rounds of self-corrections for CoSC-Code in 7B and 13B on MATH dataset.
}
\vspace{-1em}
\label{tab:the effect of the context lengths}
\small
\centering
\setlength{\tabcolsep}{18pt}{
\begin{tabular}{l|ccccc}
\toprule
Models & 0-1k & 1k-2k & 2k-3k & 3k-4k & $\textgreater$4k  \\
\cmidrule{1-6}
CoSC-Code-7B & 84.88\% & 13.68\% & 1.28\% & 0.16\% & 0  \\
CoSC-Code-13B &  85.36\% & 13.24\% & 1.30\% & 0.10\% & 0 \\
\bottomrule
\end{tabular}}
\end{minipage}
\vspace{1em}

\begin{minipage}{\linewidth}
\caption{Accuracy results (\%) on MATH and GSM8K datasets of ToRA-Code-7B, ToRA-Code-7B under CoSC prompts and our CoSC-Code-7B.}
\vspace{-1em}
\label{tab:Accuracy results on MATH and GSM8K datasets of ToRA-Code-7B, ToRA-Code-7B under CoSC prompts and our CoSC-Code-7B}
\small
\centering
\setlength{\tabcolsep}{23pt}{
\begin{tabular}{l|cc|c}
\toprule
Models & MATH & GSM8K & AVG \\
\cmidrule{1-4}
ToRA-Code-7B & 44.6 & 72.6 & 58.6 \\
ToRA-Code-7B under CoSC prompts & 42.8 & 68.0 & 55.4 \\
CoSC-Code-7B (Ours)	& 47.9 & 75.1 & 61.5 \\
\bottomrule
\end{tabular}}
\end{minipage}
\end{table}

\subsection{Alleviated Issues by our CoSC}

Our CoSC primarily targets resolving two types of issues related to (a) code errors, such as NotImplementedError, and (b) inconsistency errors between outputs and the given questions. We have quantified the reduction in errors for the 7B models on MATH dataset, as demonstrated in the results below. It can be observed that these two types of errors are effectively minimized by our CoSC method. This reduction highlights the effectiveness of our CoSC approach in enhancing the accuracy and reliability of LLMs in mathematical reasoning tasks.

(a) Statistics for error reduction related to code errors: reduced from 931 to 167.

(b) Statistics for error reduction related to inconsistency errors: reduced from 310 to 1.

\subsection{Results of ToRA under CoSC prompts}

To explore whether only using CoSC prompts can embed self-correction ability in LLMs, we conduct an experiment using the CoSC prompts for evaluation on ToRA-Code-7B. The CoSC prompts are same as those used in CoSC seeding data generation in Appendix \ref{app: Prompt for our CoSC}.  The results on the MATH and GSM8K datasets are shown in Table \ref{tab:Accuracy results on MATH and GSM8K datasets of ToRA-Code-7B, ToRA-Code-7B under CoSC prompts and our CoSC-Code-7B}.

As shown in Table \ref{tab:Accuracy results on MATH and GSM8K datasets of ToRA-Code-7B, ToRA-Code-7B under CoSC prompts and our CoSC-Code-7B}, applying CoSC prompting to ToRA not only fails to outperform the original ToRA model but also results in a decline in performance. As demonstrated in Table \ref{tab:multi step reasoning}, ToRA inherently lacks the robust multi-round reasoning capabilities needed for effective self-correction. When CoSC prompting is applied, it introduces complexity that the model is ill-equipped to handle, leading to confusion and errors in the iterative process. Similarly, during the development of the CoSC algorithm, we also attempt to apply self-correction prompts to the base CodeLLaMA model. However, this approach did not yield good performance and was significantly lower than the previous state-of-the-art results in open-source models. This led us to adopt a fine-tuning strategy instead.

In contrast, our CoSC model, which integrates self-correction as an inherent capability via fine-tuning, achieves superior results on both datasets. These findings suggest that for open-source LLMs, few-shot prompting alone is insufficient to effectively enable self-correction. The lack of significant gains from prompting further underscores the limitations of relying solely on in-context examples. Therefore, we argue that embedding self-correction as an inherent capability through fine-tuning is essential for truly endowing LLMs with robust self-correction abilities.

Moreover, by integrating self-correction directly into the training process, our approach allows models to perform self-correction autonomously in a zero-shot setting during inference, eliminating the need for external feedback or few-shot demonstrations. This self-correction mechanism enables even weaker LLMs to achieve significant improvements in mathematical reasoning—enhancements that are unattainable through prompting methods alone. Additionally, our CoSC framework is open-source, making these advancements accessible to the broader research community. We believe this represents a pivotal step toward democratizing advanced reasoning capabilities and fostering further innovation.

\section{Limitations}
\label{app: Limitations}


In this paper, we propose a novel mechanism known as the Chain of Self-Correction (CoSC) designed to enhance the reasoning capabilities of Large Language Models (LLMs). While our research primarily concentrates on mathematical reasoning, we posit that the CoSC mechanism could be effectively utilized across a wider spectrum of applications to rectify errors produced by LLMs. However, due to constraints related to the length of this paper, a comprehensive exploration of this generalization will be reserved for future study.

\section{Reproducibility Statement}
\label{app: Reproducibility Statement}
We provide part of the codes and some seeding data. We will provide the remaining codes and  data of CoSC for reproducibility upon the acceptance of the paper.




\end{document}